\documentclass{article}

 \usepackage[preprint]{amlc_2022}

\usepackage[utf8]{inputenc} 
\usepackage[T1]{fontenc}    
\usepackage{hyperref}       
\usepackage{url}            
\usepackage{booktabs}       
\usepackage{amsfonts}       
\usepackage{nicefrac}       
\usepackage{microtype}      
\usepackage{xcolor}         

\usepackage{times}
\usepackage{latexsym}
\usepackage{parsetree}
\usepackage{qtree}
\usepackage{stfloats}
\usepackage{supertabular}
\usepackage{graphicx}
\usepackage[linguistics]{forest}
\usepackage{makecell}
\usepackage{color,soul}
\usepackage{textcomp}
\usepackage{subfigure}
\definecolor{lightergray}{gray}{0.92}

\usepackage[fleqn]{amsmath}

\graphicspath{{./figures/}}

\title{PIZZA: A new benchmark for complex \\ end-to-end task-oriented parsing}

\author{%
  Konstantine ~Arkoudas \\
   Alexa AI \\
  \texttt{arkoudk@amazon.com} \\
   \And
   Nicolas ~Guenon des Mesnards \\
  Alexa AI \\
  \texttt{mesnarn@amazon.com} \\
   \AND
   \And
  Melanie ~Rubino\\
  Alexa AI \\
   \texttt{rubinome@amazon.com} \\
   \And
	Sandesh ~Swamy\\
	Alexa AI \\
	\texttt{sanswamy@amazon.com} \\
   \And
	Saarthak ~Khanna \\
	Alexa AI \\
	\texttt{saartk@amazon.com} \\
   \And
   Weiqi ~Sun \\
  Alexa AI \\
   \texttt{weiqisun@amazon.com} \\
   \And
	Khan ~Haidar\\
	Alexa AI \\
	\texttt{khhaida@amazon.com} \\
}

\newcommand{\mtt}[1]{\mbox{\tt #1}}
\newcommand{\lp}{\mbox{\tt (}}
\newcommand{\rp}{{\tt )}}
\newcommand{\egnsp}{\mbox{e.g.}}
\newcommand{\fmtt}[1]{\mbox{\footnotesize\tt #1}}
\newcommand{\smtt}[1]{\mbox{\small\tt #1}}
\newcommand{\temv}[1]{\mbox{\em #1}}
\newcommand{\exr}{\mbox{EXR} }

\begin{document}

\maketitle

%

\begin{abstract}
  Much recent work in task-oriented parsing has focused on finding a
  middle ground between flat slots and intents, which are inexpressive but easy to annotate, and powerful
  representations such as the lambda calculus, which are expressive but costly to annotate. This paper continues the exploration of task-oriented parsing by introducing a new dataset for parsing pizza and drink orders, whose semantics cannot be captured by flat slots and intents. We perform an extensive evaluation of deep-learning techniques for task-oriented parsing on this dataset, including different flavors of seq2seq systems and RNNGs. The dataset comes in two main versions, one in a recently introduced utterance-level hierarchical notation that we call TOP, and one whose targets are executable representations (EXR). We demonstrate empirically that training the parser to directly generate EXR notation not only solves the problem of entity resolution in one fell swoop and overcomes a number of expressive limitations of TOP notation, but also results in significantly greater parsing accuracy.
\end{abstract}

\section{Introduction}

Virtual assistants like Siri and Alexa are becoming ubiquitous, and their ability to understand natural language has come a long way since their early days. Traditionally such systems are based on semantic frames that assign a unique intent to every utterance and a single slot label to every  utterance token \citep{tur2011spoken,mesnil2013investigation,liu2016attention}. This results in flat representations
that are unable to capture the sort of structured semantics that are needed for even moderately complex tasks. Even something as simple as {\em I'd like two large pizzas with ham and also one diet pepsi and one coke\/} would be challenging to represent with a single intent and flat slots.\footnote{For example, it is not sufficient to merely tag the numbers; they must be properly grouped with the corresponding orders.
}

\begin{figure*}[h!]
    \centering
    \includegraphics[width=0.8\textwidth]{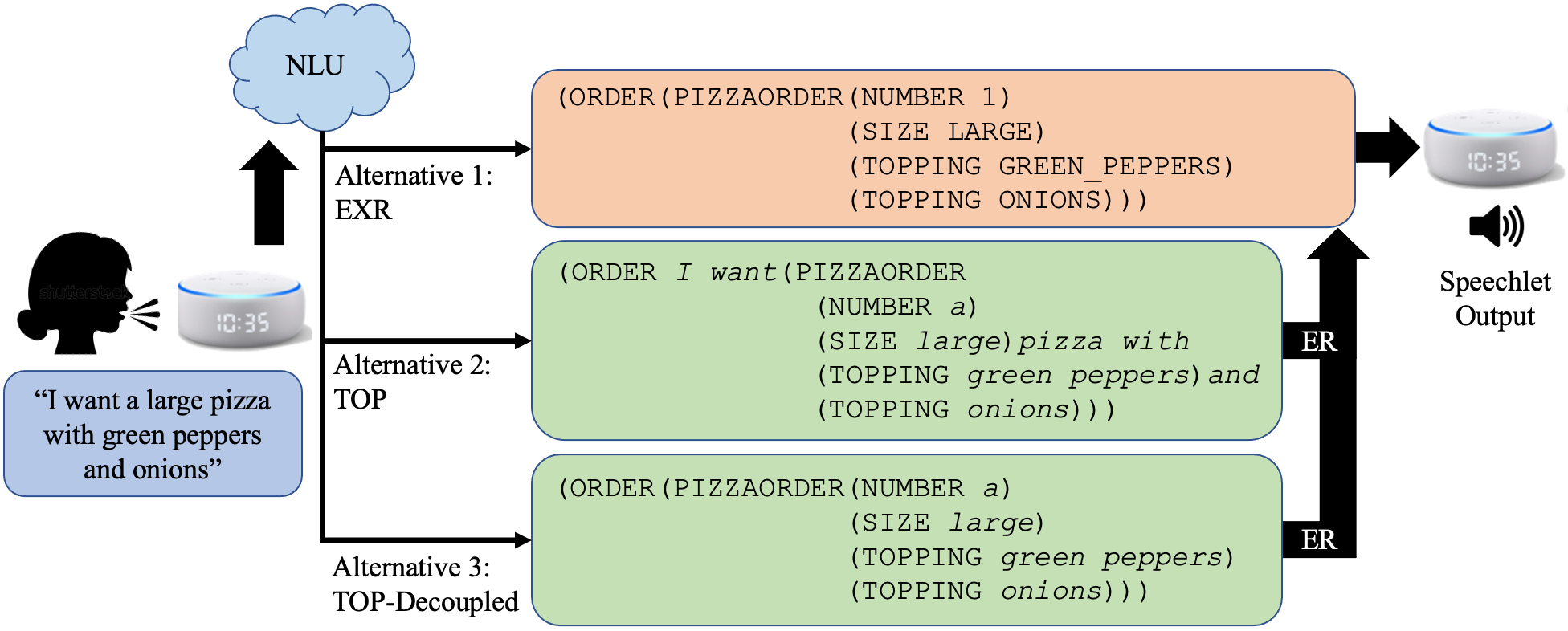}
    \caption{End-to-end task-oriented parsing workflows when using EXR, TOP, and TOP-Decoupled representations.}
    \label{fig:parsing_schematic}
\end{figure*}

The semantic parsing community has a long history of exploring very rich semantic representations, such as the typed lambda calculus and dependency graphs \citep{zettlemoyer2012learning,banarescu2013abstract,berant2014semantic}. However, annotating utterances with such intricately detailed representations is difficult, so these have not found wide adoption in the industry. Recent work on task-oriented parsing has sought to increase expressivity without imposing an excessive burden on annotation or parsing. A prominent thread of work here was launched by the introduction of a hierarchical notation that we will call TOP\footnote{The term \fmtt{TOP} is also widely used to refer to a particular dataset that was introduced by \cite{gupta2018semantic}. Here we use the term to refer to the general representational scheme rather than that specific dataset.} in which the utterance text and its semantics are organized into a tree structure similar to a constituency parse tree. When linearizing such a tree, pairs of matching parentheses are used to achieve the necessary nesting. For instance, annotating the foregoing example might result
in the following string:

 {\footnotesize
\begin{tabular}{l}    
\hspace*{-0.15in}  \lp{\tt ORDER} {\em I'd like} \lp{\tt PIZZAORDER} {\tt (NUMBER} {\em two}\rp  {\tt (SIZE} {\em large\/}\rp {\em pizzas with\/} {\tt (TOPPING} {\em ham\/}\rp\rp  {\em and also\/} \\ 
\hspace*{-0.15in}  {\tt (DRINKORDER} {\tt (NUMBER} {\em one}\rp {\tt (DRINKTYPE}   {\em diet pepsi}\rp\rp {\em and} {\tt (DRINKORDER} {\tt (NUMBER} {\em one}\rp \\
\hspace*{-0.15in}    {\tt (DRINKTYPE} {\em coke}{\tt )))}
\end{tabular}}

This approach injects semantics into the utterance text by hierarchically wrapping semantic constructors---such as \smtt{PIZZAORDER}---around appropriate utterance segments. These constructors can be viewed as composable slots and/or intents. The leaves of this tree, read from left to right, reconstruct the full utterance text.

This annotation scheme has expressive limitations of its own, e.g., it cannot accommodate utterances with non-projective semantics, such as {\em I would like three pizzas, one with peppers and the others with ham}, which could be done with more expressive representations. 
Nevertheless, TOP notation is much more expressive than flat intents and slots and considerably more accessible to annotators than heavier-duty formalisms, so it presents a reasonable compromise between expressivity and practicality. In addition, it has been shown that deep-learning models can be taught to map utterances to this type of notation effectively \cite{gupta2018semantic}. 


Ultimately, semantic parsing should be producing {\em executable representations\/} (\exr for short) that a back end service can process directly. An executable representation for the above pizza order would be something along the following lines: \\[0.05in]
\begin{minipage}{3in}
{\footnotesize  
\begin{verbatim}
(ORDER (PIZZAORDER (NUMBER 2) 
                   (SIZE LARGE)
                   (TOPPING HAM))
       (DRINKORDER (NUMBER 1) 
                   (DRINKTYPE DIET_PEPSI))
       (DRINKORDER (NUMBER 1) 
                   (DRINKTYPE COKE)))
\end{verbatim}} 
\end{minipage}\\[0.05in]
Note that this representation does not contain any natural language text, unlike the earlier TOP representation. 



As part of this paper we propose a discussion on the advantages and drawbacks of using either of those representations in a production environment such as that of Alexa. We also provide an empirical study to quantify and analyze the effectiveness of models to produce such representations.

If the semantic parser outputs TOP notation, then a separate entity-resolution stage must be implemented, along with integration logic for producing the final EXR. Figure~\ref{fig:parsing_schematic} contrasts the end-to-end workflow of a parser that produces EXR vs that of one that produces TOP (the third alternative, TOP-Decoupled, is a variant of TOP we will discuss in the next section). 
Training a semantic parser to generate \exr directly obviates the need to maintain a separate downstream entity resolution (ER) component and annotating training data into \exr format would not incur extra burden. 


The EXR approach is able to handle non-projective and other challenging types of semantics, as long as these are expressible in EXR notation. To take the preceding example of {\em three pizzas, one with peppers and the others with ham}, the target EXR semantics would simply be the conjunction of two clauses of the form \smtt{(PIZZAORDER} \smtt{(NUMBER 1)} \smtt{(TOPPING PEPPERS))} and \smtt{(PIZZAORDER} \smtt{(NUMBER 2)} \smtt{(TOPPING HAM))}. Constructs of this form can be easily generated by probabilistic context-free grammars (PCFGs) with semantic actions (more on these below), and potentially learned by statistical models like DNNs (Deep Neural Networks).


To be able to measure the effectiveness of NLU models to predict either of those more expressive target representations, we release PIZZA, a new task-oriented parsing benchmark. Unlike the TOP dataset from \cite{gupta2018semantic}, PIZZA allows to evaluate systems end-to-end by providing EXR representations. The training set of PIZZA consists of a large collection of synthetically generated utterances, while the dev and test sets are much smaller and human-generated.
The dataset comes in three versions, one in EXR notation, one in TOP notation, and one in TOP-Decoupled (discussed in next section).



Oftentimes the NLU system is bootstrapped with a grammar $G$ that emits semantic representations. That grammar is then sampled to produce utterances that serve as training data for a statistical model $M$. At runtime, an utterance $u$ is analyzed by running $G$ and $M$ concurrently and their results are then combined by some algorithm (e.g., merged and reranked), with the results of $G$ often taking precedence, particularly when $G$ parses $u$ successfully and only produces one interpretation.
This is also the scenario that we emulate in this paper. Because intents and slots are typically flat, $G$ is usually a finite state transducer (FST) that outputs an intent and a label for each utterance token. But since FSTs are unable to capture scope and unbounded stack-based recursion, we instead use a hand-crafted PCFG augmented with semantic actions that directly produce EXR. We used the derivations produced by this grammar to automatically produce TOP counterparts of the EXR training data, and likewise for the dev and test sets, though for dev and test utterances that were not parsable by the PCFG, TOP annotations had to be given by hand.

Our contributions in  this paper are as follows: 
\begin{enumerate}
	
	\item We release a new dataset\footnote{\url{https://github.com/amazon-research/pizza-semantic-parsing-dataset}} for task-oriented parsing, in three versions,  EXR, TOP, and TOP-Decoupled. 
	
	\item We perform an extensive empirical evaluation of different deep-learning techniques for task-oriented parsing, including a number of seq2seq architectures, Recurrent Neural Network Grammars (RNNGs), and a new pipeline technique that works by flattening tree structures.
	
	
	\item We show that systems that generate EXR notation significantly outperform their
	TOP-generating counterparts, which calls for further research in generating
	executable semantic representations with resolved entities instead of a blend of utterance tokens and semantic constructors.
	
	\item While the PCFG achieves remarkable accuracy on its own (68\%),  we show that seq2seq systems generalize beyond the PCFG that was used to train them. Specifically, the best performing seq2seq system parses 60\% of the utterances that the PCFG cannot handle.

\end{enumerate}

\section{The PIZZA dataset}
\label{Sec:Dataset}
The main characteristics of the natural-language orders comprising the PIZZA dataset can be summarized as follows. 
  Each order can include any number of pizza and/or drink suborders. These suborders are labeled with the constructors {\small PIZZAORDER} and {\small DRINKORDER}, respectively.
  Each top-level order is always labeled with the root constructor {\small ORDER}. 
  Both pizza and drink orders can have \textsc{number} and \textsc{size} attributes. 
  A pizza order can have any number of \textsc{topping} attributes, each of which can be negated. Negative particles can have larger scope with the use of the \temv{or} particle, e.g., {\em no peppers or onions\/} will negate both peppers and onions. Toppings can be modified by quantifiers such as {\em a lot\/} or {\em extra}, {\em a little\/}, etc. 
  A pizza order can have a \textsc{style} attribute (e.g., \textit{thin crust} style or {\em chicago\/} style).
  Styles can be negated.
  Each drink order must have a \textsc{drinktype} (\egnsp, \textit{coke)}, and can also have a \textsc{containertype} (\egnsp, \textit{bottle}) and/or a \textsc{volume} modifier (\egnsp,  {\em three 20 fl ounce coke cans\/}). We view {\small ORDER}, {\small PIZZAORDER}, and {\small DRINKORDER} as intents, and the rest of the semantic constructors as composite slots, with the exception of the leaf constructors, which are viewed as entities (resolved slot values).

\begin{figure}[ht!]
	\scriptsize{
	\centerline{ 
	\subfigure[] {
		\begin{forest}
			[ORDER [PIZZAORDER [NUMBER [1]] [SIZE [MEDIUM]] [TOPPING [PEPPERS]] [TOPPING [HAM]] [NOT [TOPPING [ONIONS]]]]]
		\end{forest}}\hspace{1cm}%
	\subfigure[] {
		\begin{forest}
			[ORDER [PIZZAORDER [NUMBER [\temv{one}]] [SIZE [\temv{medium-size}]] [\temv{pizza with}] [TOPPING [\temv{peppers}]] [\temv{and}] [TOPPING [\temv{ham}]] [\temv{but no}] [NOT [TOPPING [\temv{onions}]]]]]
		\end{forest}}
	}
}

	\caption{(a) EXR and (b) TOP representation, for the order \textit{one medium-size pizza with peppers and ham but no onions}.}
	\label{fig:examplesEXRTOP}
\end{figure}
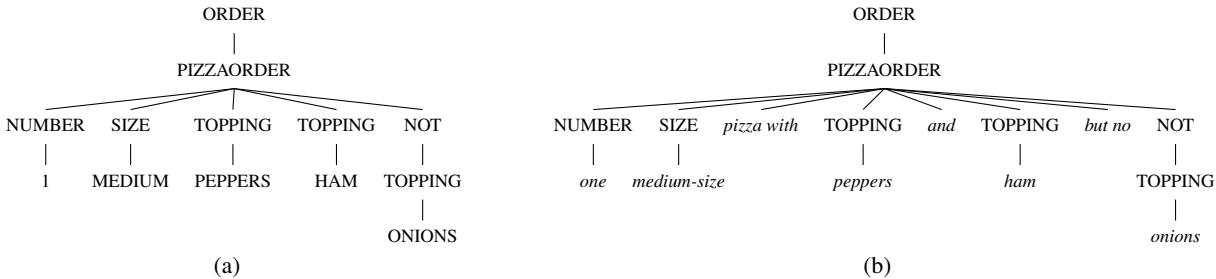

A simple example of an order is the query {\em one medium-size pizza with peppers and ham but no onions}. Figure~\ref{fig:examplesEXRTOP} (a) depicts an EXR semantic tree for this order. 
Note that while the order of siblings in EXR trees has no effect on operational semantics (more precisely, every internal tree node is commutative and associative), using a consistent ordering scheme can have a nontrivial impact on learnability (see Appendix~\ref{App:PermExps}).

The tree in Figure~\ref{fig:examplesEXRTOP} (b) is a TOP representation of the same utterance. Here, internal tree nodes are {\em not\/} commutative or associative, because permuting two subtrees will violate the constraint that a left-to-right traversal of the leaves must reconstruct the utterance text. It has been recognized before that this constraint imposes expressive limitations on TOP representations. In particular, TOP is not well-positioned to handle long-range dependencies. An example given by~\cite{conversationalTOP} is the utterance \mbox{\em On Monday, set an alarm for 8 AM}. It would be preferable here for the representation to have a single date-time constructor (slot) containing both the day and the time, but in TOP this is impossible due to the intervening text {\em set an alarm\/} separating {\em Monday\/} from {\em 8 AM}.
While these limitations may be of relatively little practical importance in English, they would be more acutely felt in languages with more flexible word orders. 
\cite{conversationalTOP} mitigate this by pruning tokens that do not appear as children of leaf slots. They call the resulting notation {\em decoupled TOP}. In this example, the decoupled TOP tree would be identical to that shown in Figure~\ref{fig:examplesEXRTOP} (b), except that the leaves corresponding to the utterance segments {\em pizza with}, {\em and}, and {\em but no\/} would be removed.

The training data of PIZZA was synthetically generated by sampling a subset of our grammar (see Appendix \ref{app:gram}).
Because the grammar was written to maximize recall, only a relatively small subset of its rules were sampled.
Care was taken to generate natural-sounding orders and to avoid pathological instances, such as {\em a pizza with mushrooms and no mushrooms}.
Once the utterances were generated, their EXR semantics were obtained by running our parser on them (the grammar generates EXR directly).
TOP counterparts were automatically generated by an algorithm that analyzes the EXR derivations produced by the parser.
The total number of synthetically generated utterances are about 2 million, but because the number of sampled patterns is small,
the training set has more lexical than structural diversity (as is often the case with the initial data of grammar-based systems facing a cold start).
Consequently, the dataset has a good amount of structural redundancy and a well-pretrained model fine-tuned on a small subset of the
data can achieve performance on par with a model trained on the full dataset (see the ablation results in Appendix~\ref{app:ablation}).




The test and validation data are human-generated and were collected by two MechanicalTurk tasks: (1) A paraphrasing task where the annotators were shown a
synthetically generated order in plain English and were asked to rephrase it in a more natural way.
(2) A text generation task where annotators were asked to formulate orders of their choice, for themselves or for a group of people.
This free-style collection process resulted in 1K unbiased natural language orders, from which we manually extracted the semantics in \mbox{EXR} format.
Each utterance was annotated by two individuals, and we only kept the ones where both annotations matched.
The \mbox{TOP} versions of the dev and test sets were obtained by first running our parser on the utterances, analyzing the derivations to
extract the necessary TOP information, and then manually correcting the output trees as needed.
\phantomsection\label{top2topsdecoupled}
Finally, the dev and test portions of \mbox{TOP-Decoupled} were obtained from \mbox{TOP} by removing tokens that do not appear as children of leaf slot constructors.
Additional dataset creation details can be found in Appendix~\ref{App:DataAndModelDetails}, and dataset examples in Appendix~\ref{app:ex}.

The following table presents some statistics on the dataset:  
{\small
\begin{center}
	\begin{tabular}{|l||c|c|c|}
		\hline
		& \makecell{Train} & \makecell{Dev} & \makecell{Test} \\	\hline
		Number of utterances	   		&  2,456,446  &	 348  & 1,357  \\	\hline
        Unique entities     & 367 & 109 & 180 \\ \hline

		PCFG accuracy		   	   &  100\% 	& 70\% & 68\%	\\	\hline
		\begin{tabular}{@{}c@{}}Avg entities per utterance\end{tabular}		   	   &  5.32  & 5.37	 & 5.42	 \\	\hline
		\begin{tabular}{@{}c@{}}Avg intents per utterance\end{tabular}		   	   &  1.76  & 1.25	 & 1.28	 \\	\hline
	\end{tabular}
\end{center}} \noindent 
A significant proportion of human-generated utterances requires generalization from synthetic data and cannot be fully parsed by the PCFG parser.

While the domain of pizza ordering is conceptually simple (and familiar to all), the semantics can get fairly subtle. For instance, in the query {\em I'd like four pizzas, half with ham and half with peppers}, it's not obvious if the request is for two pizzas with ham and two with peppers, or for four pizzas where the half of each has peppers and the other half ham.
Another factor making this particular dataset challenging is the distribution shift between the training and testing data. This is also a common theme in practice, as semantic parsing systems are usually faced with a cold start problem, whereby initial versions are built by developers who may not be able to anticipate user queries in the wild. Hence, test utterances collected subsequently may diverge significantly from the utterances that the initial system was designed to handle.


\section{Models}
\label{Sec:Models}

In this section we introduce seq2seq models as well as a new pipelined approach. We also experimented with Insertion Transformers~\cite{stern2019insertion} and RNNGs~\cite{dyer2016recurrent} but defer the details to Appendix~\ref{app:app_other_models}.

\subsection{Seq2seq architectures}

\begin{figure*}[h!]
\centering
        \includegraphics[width=0.9\textwidth]{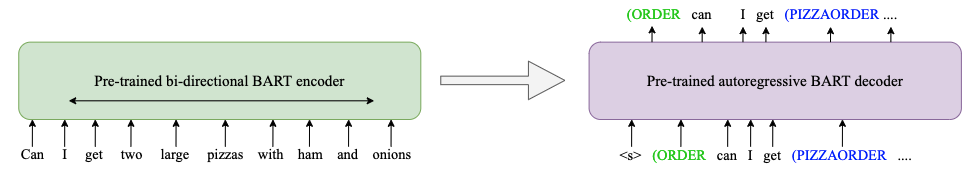}
        \caption{Our Transformer-based seq2seq architecture.}
    \label{fig:bart_s2s}
\end{figure*}

We model the generation of EXR/TOP representations from natural
 language utterances as a sequence-to-sequence (seq2seq) learning problem
 \cite{sutskever2014} where the source sequence is the 
 utterance and the target sequence is a linearized rendering of the EXR/TOP tree (obtained from a preorder traversal of the tree).
 For encoding and decoding we explored Long-Short Term Memory (LSTMs) 
 \cite{schmidhuber-97} as well as Transformer encoders \cite{vaswani17} and BART decoders \cite{lewis2019bart}
 as the starting  point for our task. Figure \ref{fig:bart_s2s} illustrates this setup
 when  using a Transformer-based encoder and decoder; the target sequence is a TOP 
 representation. 

The decoder is fine-tuned (trained from scratch in case of LSTMs) to generate 
both natural language tokens and semantic constructors like \texttt{PIZZAORDER}.
For the LSTMs we used a single-layer 512-dimensional bi-directional encoder and 
decoder with attention and pretrained fastText embeddings \cite{bojanowski2017enriching}. For the Transformers, we performed experiments with the Large (12-layer) variation of BART. 
We kept the BART token and position 
embeddings frozen and only unfroze the other layers. We do not constrain the decoder vocabulary and preserve the original pre-trained vocabulary. We train our models to minimize the sequence cross-entropy loss against the chosen target representation.

\subsection{Pipeline model}
We also introduce a ``divide, flatten, and conquer'' approach that uses a pipeline of two models run in sequence:
an intent segmentation (IS) model to determine the intent spans present in the input utterance, and a conventional 
named entity recognition (NER) model to assign flattened entity labels to the tokens of each intent span identified by the first model.
These  models are trained separately on two different sequence labeling tasks. For example in {\em two large pizzas with ham and one diet coke\/}, the labeling for 
the intent segmentation model would be: \\[0.05in]
\begin{minipage}{3in}
{\footnotesize  
\begin{verbatim}
B-PIZZAORDER I-PIZZAORDER I-PIZZAORDER 
I-PIZZAORDER I-PIZZAORDER Other 
B-DRINKORDER I-DRINKORDER I-DRINKORDER.
\end{verbatim}} 
\end{minipage}\\[0.05in] 
The example carries two different intent spans, one for a pizza order and one for 
a drink order. As a result, the NER model will have two different inputs: {\em two 
large pizzas with ham\/} and {\em one diet coke\/}. The NER labels for each  will be: 
\begin{minipage}{3in}
{\footnotesize  
\begin{verbatim}
B-NUMBER B-SIZE Other Other B-TOPPING
\end{verbatim}} 
\end{minipage}\\[0.05in] 
and \begin{minipage}{3in}
{\footnotesize  
\begin{verbatim}
B-NUMBER B-DRINKTYPE I-DRINKTYPE.
\end{verbatim}} 
\end{minipage}\\[0.05in] To handle cases where hierarchy is needed, 
we compress more information into flattened labels, e.g., a negated 
pizza topping will be labeled as \smtt{NEG\_TOPPING}. Because these models need to have a 1:1 mapping between the input tokens and output labels, they can't be trained on the EXR notation or the TOP-Decoupled notation. Both models use a BERT-based  encoder \cite{devlin2019bert}. 


\section{Results and analysis}\label{sec:results}
The main results\footnote{In Appendix~\ref{app:app_other_models} we provide results for other models.} are summarized in Table \ref{table:exrres}.
EXR is taken as the ground truth, so the outputs of models that produce TOP or TOP-Decoupled notation must then undergo
entity resolution (ER) in order to produce a final output in EXR format, which can be compared against the ground truth.
See Appendix \ref{app:topperfectER} for ER details.

Our metric is Exact Match accuracy \textbf{(EM)}, which checks for an exact match between the ground truth and prediction
semantic trees, modulo sibling order. In addition, for the models that produce TOP representations, we throw out all
utterance tokens that are not in a leaf slot.

\begin{center}
	\begin{table*}[!hbt]
		{\small
			\centering
			\caption{EM against EXR ground truth, grouped by notation type used for training. Mean and standard error from
				models trained on 5 different seeds.}
			\smallskip
			\begin{center}
				\begin{tabular}[c]{lccc}
					\hline
					\textbf{Model} & \thead{\textbf{Dev} \\ \textbf{(348 utt)}} & \thead{\textbf{Test} \\ \textbf{(1357 utt)}}  &  \thead{\textbf{PCFG Error} \\ \textbf{Test Subset} \\ \textbf{(434 utt)}} \\
					\hline
					PCFG Parser                           & 69.54 & 68.02 & 0.0  \\ \hline
					\multicolumn{1}{l}{\em TOP\/}         & & & \\
					\hspace{3mm}LSTM                              & 44.42 \texttt{\textpm} 2.02 & 41.44 \texttt{\textpm} 2.15 & 12.63 \texttt{\textpm} 0.50 \\
					\hspace{3mm}BART                              & 63.91 \texttt{\textpm} 0.31 & 62.17 \texttt{\textpm} 0.31 & 27.56 \texttt{\textpm} 0.31 \\
					\hspace{3mm}Pipeline                          & 68.10 \texttt{\textpm} 0.44 & 64.08 \texttt{\textpm} 0.32 & 20.97 \texttt{\textpm} 0.96 \\
					\multicolumn{1}{l}{\em TOP-Decoupled\/}& & & \\
					\hspace{3mm}LSTM                              & 46.55 \texttt{\textpm} 1.03 & 42.15 \texttt{\textpm} 1.52 & 14.79 \texttt{\textpm} 1.11 \\
					\hspace{3mm}BART                              & 74.60 \texttt{\textpm} 0.15 & 71.73 \texttt{\textpm} 0.13 & 40.55 \texttt{\textpm} 0.50 \\
					\multicolumn{1}{l}{\em EXR\/}         & & & \\
					\hspace{3mm}LSTM                              & 38.51 \texttt{\textpm} 3.58 & 33.28 \texttt{\textpm} 4.17 & 14.93 \texttt{\textpm} 1.22 \\
					\hspace{3mm}\textbf{BART}   &  \textbf{81.26 \textpm$\:$0.48} & \textbf{78.56 \textpm$\:$0.31} & \textbf{59.49} \textpm$\:$\textbf{0.74} \\
					\hline
				\end{tabular}
			\end{center}
		\label{table:exrres}
		
	}
\end{table*}
\end{center}

\paragraph{Generating EXR improves performance.} Training BART to generate TOP-Decoupled instead of TOP improves performance from 62\% to 72\% EM. Generating EXR further improves BART model by 7-EM points.\footnote{LSTM results do not show the same trend for EXR, but the results are merely given as a baseline for which no extensive hyperparameter search was carried out.} 
This change not only gives better performance, owing to a smaller decoder vocabulary and shorter output sequences, but also generates
the final executable representation needed for downstream processing by the voice assistant. 
%
%

\begin{figure}[h!]
	\centerline{
		\subfigure[]{ \includegraphics[width=0.44\textwidth]{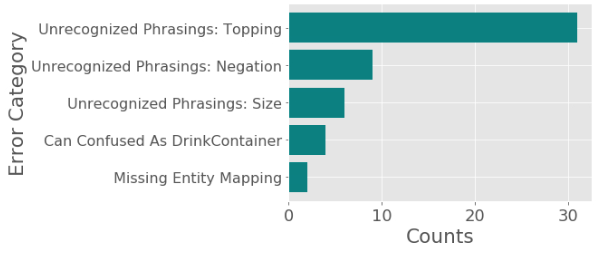} }
		\subfigure[]{  \includegraphics[width=0.38\textwidth]{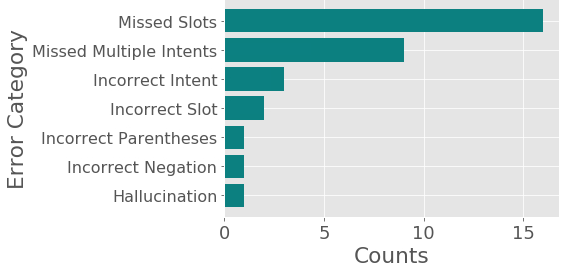} }
		\subfigure[]{ \includegraphics[width=0.38\textwidth]{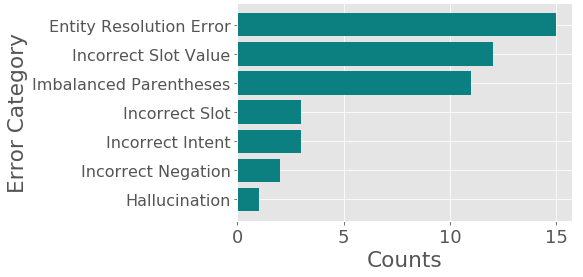} } }
	\caption{(a) PCFG dev set errors that BART EXR got correct (b) BART EXR dev set errors that PCFG got correct (c) BART TOP-Decoupled dev set errors that BART EXR got correct}
	\label{fig:errors}
\end{figure}

\paragraph{Generating EXR adds more than just entity resolution.}
One might consider the poor performance of TOP or TOP-Decoupled compared to EXR is strictly due
to the entity resolution step.
However, in manual error analysis\footnote{Manual error analysis was performed on the
	dev data set, where a single example may be counted in multiple error categories.} of BART models, shown in Figure \ref{fig:errors} (c),
only 30\% of TOP-Decoupled errors that BART on EXR got correct could have been improved with perfect entity resolution, though many of these examples also
had other errors in them.


In addition, Appendix \ref{app:topperfectER} presents similar EM results on parsing only, without subsequent ER, and also when perfect ER is applied, further confirming that the ER step is not the main reason for the poorer performance of these models.

\paragraph{Generating EXR comes with challenges.} While the gains in performance in generating EXR call for further research on end-to-end systems, it must be noted that the integration of such models in production has some caveats. Notably, every time we want to expand the catalogs with new entities, the whole system needs to be retrained to augment its output space. That being said, it would likely not be retrained from scratch and only a couple model updates could suffice. For a pipelined system, the ER component would also need to be retrained, but the NLU component could remain unchanged. It is important to note that generating EXR is more appealing for a closed-world task like food-ordering where there is a finite menu with likely less than a couple thousand entities. For applications such a Music where there are dozens of millions of artist and song names, directly generating EXR brings additional challenges that need to be addressed through further innovations.

\paragraph{Some seq2seq models outperform the PCFG.}
The PCFG parser achieves high accuracy on its own (68\%), showing that a well-crafted grammar can be highly effective by itself. 
However, powerful pretrained language models and judicious choice of target notation and architecture can yield DNN models that generalize
beyond the PCFG, even though they were trained only on PCFG-generated data.  In particular, BART trained on TOP-Decoupled and EXR achieves 72\% and 79\% EM, respectively.
In addition, on the portion of the test set that PCFG gets wrong, called \emph{"PCFG Error Test Subset"} above, BART on EXR reaches 60\% EM.

\paragraph{Seq2seq complements the PCFG.}
Most of the PCFG errors in the dev set that BART got correct were due to diverse phrasings used to specify toppings, sizes and
negations.  Although many common phrasings are captured in the grammar, not all can be handled a priori.  This is where
a seq2seq model fine-tuned on a strong language model can outperform.  Consider the request {\em i'd like a large pizza
	with sausage and ham and also add some extra cheese}. It is challenging for the original PCFG to parse {\em extra cheese}
as a topping given the preceding phrase {\em and also add some}, but  BART trained on EXR makes the correct prediction in this case. 
Manual analysis of 50 out of the 70 dev set errors in Figure~\ref{fig:errors} (a) shows that BART on EXR generalizes beyond
the phrasing patterns defined by the PCFG.\footnote{Semi-supervised learning from live traffic could be used to improve PCFG structural diversity.}.

On the other hand, there are commonly phrased requests that the PCFG correctly parses but BART gets wrong; see Figure
\ref{fig:errors} (b) summarizing those 31 errors, 9\% of dev set.  In lengthier requests, especially those with multiple intents,
BART trained on EXR misses slots and doesn't generate all of the intents requested.  Consider the request {\em i would like to order a medium pizza with
	italian sausage a small pizza with beef and mushrooms a large combination pizza and four large pepsis}, which the 
PCFG correctly parses.  While BART on EXR
correctly generates a \texttt{PIZZAORDER} intent for the medium pizza and the \texttt{DRINKORDER} intent, the other two
\texttt{PIZZAORDER} intents get incorrectly combined into one with two \texttt{TOPPING} slots missing.

The pipeline model does very well (64\%), but seems less complementary with the PCFG than BART. This model would also not be an appropriate choice for a domain with deep semantic trees.

\section{Related Work}
\label{Sec:RelatedWork}

Semantic parsing is traditionally conceived as mapping natural language to logical forms that can be either directly executed or readily translated into an executable form, and the community has a long history of studying richly expressive languages for such logical forms, most notably variants of the typed lambda calculus \cite{ZelleM96,zettlemoyer2012learning}. While these representations can capture the semantics of a tremendous range of natural language constructs,  they are challenging to  annotate and to parse. There is also a long tradition of shallow semantic representations, particularly in the context of task-oriented parsing, using semantic frames based on intents and slots; prominent datasets here have been ATIS \cite{ATIS} and more recently, SNIPS \cite{coucke2018snips}. These representations are much more limited in their expressivity, but they are much more accessible to annotators and easier to parse. 

Recent work in the field has focused on increasing the expressive power of task-oriented parsing, moving beyond flat intents and slots to accommodate compositionality, but without overly complicating annotation and parsing. \cite{gupta2018semantic} introduced the ``TOP notation'', a tree-based representation for task-oriented parsing with the key property that traversing the leaves from left to right reconstructs the utterance text. More recent work has introduced a so-called ``decoupled'' variant of TOP \cite{conversationalTOP}.

\section{Conclusions}
\label{Sec:Conclusions}

We have introduced a new dataset for task-oriented parsing, PIZZA, with three notational versions, EXR, TOP, and decoupled TOP. EXR semantic trees are variable-free, directly executable by the back end, and contain no utterance tokens, unlike both TOP and decoupled TOP. 
We performed an extensive empirical evaluation of a number of DNN-based techniques for semantic parsing on this dataset.

A key original motivation for introducing TOP notation was that its structure is similar to standard constituency parses, allowing for easy adaptation of algorithms developed for phrase structure parsing for inference, algorithms such as linear-time RNNGs. This does not appear to be a compelling consideration at present, as more recent results have shown that direct seq2seq approaches based on transformer architectures and large pretrained models outperform techniques such as RNNGs. That is consistent with the findings presented in this paper.

\begin{ack}
The authors thank Chi-Liang Liu for his diligent work in the early stages of this project, and Saleh Soltan for helping us to use BART models in our framework. We thank Stephen Rawls for his careful review and are also indebted to all the colleagues who helped in the review, and with manual inspection and annotation of hundreds of utterances.
\end{ack}

\clearpage

{
\small

\bibliography{anthology,custom}
\bibliographystyle{acl_natbib}

}

\clearpage
\appendix
\section{PCFG}\label{app:gram}

In our PCFG framework, non-terminals are executable machines akin to stochastic RTNS (recursive transition networks) or parser combinators \cite{Fokker95}. To write a production for such a non-terminal is to write declarative code for the corresponding machine. This code is written in a custom-designed functional programming language that provides special-purpose facilities for defining and parsing grammars with arbitrarily complicated semantic actions. As an example, the following code snippet defines a  grammar for the paradigmatic context-free language of balanced parentheses $\{a^n \, b^n \: | \: n \geq 0\}$:
\begin{verbatim}
def S = id + "a" * S * "b"
\end{verbatim}
The addition sign \mtt{+} indicates alternation (akin to the pipe \mtt{|} of the more conventional textbook notation) and the multiplication sign \mtt{*} indicates juxtaposition (just white space in conventional notation); the latter binds tighter than the former. The keyword \mtt{id} stands for the $\epsilon$ (empty) string of formal language theory, and quoted strings act as terminals. The entire input is a definition (indicated by \mtt{def}) that binds the name \mtt{S} to the corresponding recursive machine (non-terminal).

Semantic actions can be inserted anywhere in a machine definition and manipulate an implicit {\em semantic stack\/}, an abstraction that is baked into the operational semantics of this programming language. This is a stack of EXR trees built during semantic parsing. Basic data types such as numbers, strings, and nullary semantic constructors (constants) serve as tree leaves. For instance, suppose we view the semantics of a string of $n$ balanced parentheses pairs $a^n \, b^n$ as the number $n$. Then here is how we might augment the above grammar with appropriate semantic actions to extract that number and deposit it on the semantic stack upon conclusion of parsing: 
\begin{verbatim}
def push(t) = fun S => t::S
def succ = fun n::S => n+1::S
def S = push(0) + 
        "a" * S * "b" * succ
\end{verbatim}
Here \mtt{push} is a semantic action that pushes an arbitrary term (tree) on the semantic stack, and \mtt{succ} is another action that increments the number that is assumed to be on the top of the stack. The syntax form \mtt{fun} $x_1,\ldots, x_n$ \mtt{=>} $e$, where $e$ is an arbitrary expression of this language, defines an anonymous higher-order function. A stack is just a list that grows on its left end, and \mtt{::} is a list constructor (so that $x\mtt{::}l$ denotes the list obtained by prepending $x$ to the list $l$).

The language allows for direct use of slot catalogs inside machine definitions and scales to catalogs with millions of entries.
These catalogs are like slot value gazetteers, except that they map phrases to unique entity identifiers, thus enabling entity resolution to be bundled up with semantic parsing. A catalog entry is therefore a triple of the form $(t_1 \cdots t_n,e,p)$ where $t_1 \cdots t_n$ is a phrase consisting of one or more tokens, $e$ is a globally unique identifier denoting an entity, and $p$ is the conditional probability of the given phrase denoting the given entity, conditioned on the catalog. The phrase $t_1 \cdots t_n$ is called an entity {\em alias}.

\begin{table*}[h!]
  \centering  
{\small 
	\caption{Statistics on the entity aliases used in the definition of the pizza grammar. Each entity, like \smtt{PEPPERS}, can have more than one natural language alias, for example \smtt{peppers} or \smtt{bell peppers.}} \label{table:tries}	
	\smallskip
	\begin{tabular}{|l||c|c|}
		\hline
		 Slot catalog &  \makecell{Number of unique slot entities} & \makecell{Average number of \\ aliases  per entity} \\	\hline
		\smtt{TOPPING} & 85 & 1.88  \\ \hline
		\smtt{SIZE} & 8 & 2.25  \\ \hline
	    \smtt{NUMBER} &  15  & 2.27    \\ \hline
	    \smtt{STYLE} &  23   & 2.39    \\ \hline
	    \smtt{QUANTITY} &  2  & 18.5    \\ \hline
	    \smtt{DRINKTYPE} &  22  & 2.77    \\ \hline
	    \smtt{VOLUME} &  11  & 9.18    \\ \hline
	    \smtt{CONTAINERTYPE} &  2  & 8    \\ \hline
\end{tabular}}
\end{table*}

The pizza domain has only a few small slot catalogs. Some basic statistics for these are shown in Table~\ref{table:tries}. As mentioned above, each entity may be associated with multiple aliases, each with its own conditional probability. For instance, {\em ricotta\/} and {\em ricotta cheese\/} may be two different aliases for the \smtt{TOPPING} entity \smtt{RICOTTA\_CHEESE}.

The main parsing technique of this framework is a probabilistic top-down algorithm with memoization that is written as an inference system for an abstract execution machine. Bottom-up parsing is also implemented to allow for left recursion, but top-down parsing is typically more efficient. The pizza grammar is written in this language in about 50 lines of code, plus the catalogs. Left recursion is not needed, so utterances are parsed with the top-down algorithm.

The algorithm takes an utterance $u$ as input and produces a ranked list of triples $(e_1,d_1,p_1)\ldots,(e_n,d_n,p_n)$ as output, where each $e_i$ is an EXR, $d_i$ is a formal derivation tree that details the incremental construction of $e_i$, and $p_i$ is the derivation's probability. The list is ranked by derivation probability, so that the first EXR, $e_1$, provides the most likely interpretation of $u$. Probabilities are initially set by MLE (Maximum Likelihood Estimation) without a prior. Even in that simple setting, we have found PCFG derivation probabilities to be useful, as they result in a bias for shorter derivations, which are simpler explanations of sorts, essentially enforcing a form of Occam's razor. A subset of this grammar was sampled to produce the training utterances.

\section{Dataset details}
\label{App:DataAndModelDetails}
We manually reviewed each paraphrase and kept 1,264 correct paraphrases. Each utterance was paraphrased by five different annotators. We found that 6\% of the paraphrases were incorrect; most omitted one topping or the relevant quantity in a suborder. See Table~\ref{table:expara} for examples.

\begin{table*}[h!]
  \centering
{\small   
	\caption{Subset of utterances obtained from the Mechanical Turk paraphrasing task.} \label{table:expara}
	\smallskip
	\begin{tabular}{|l||l|c|}
		\hline
		\makecell[l]{Synthetically \\ generated order} & Obtained paraphrase & Valid\\	\hline
		&I think I'll have a small thin crust pie with tuna and sausage.  & Yes \\
		&		\makecell[l]{Can I try the small sausage and tuna pie and I would \\ like thin crust please.} & Yes \\
		\makecell[l]{small pie with sausage and \\ tuna and thin crust please}  &	\makecell[l]{i only want a small pizza and please for toppings i'll take \\ sausage and tuna and i'd like thin crust. thank you} & Yes \\
		& Order a small thin crust pizza with sausage and tuna.  & Yes \\
		&	I need a small sausage and tuna pizza with a thin crust. & Yes \\ \hline
		& I need five large fantas, two pepsis and one coke please. & Yes \\
		& Get me two Pepsis, a Coke, and five large Fantas. & Yes \\
		\makecell[l]{two pepsis and a coke and \\ five large fantas} & \makecell[l]{i'm here for some soft drinks. i need a \textbf{couple} of pepsis, \\ one coke, and i'll take a total of five \\ fantas; make them large. that's all.}. & \textbf{No} \\
		&i'd love to have two pepsis and a coke along with five large fantas & Yes \\
		&	Can I get two pepsis, five large fantas and a coke.. & Yes\\\hline

	\end{tabular}
        }
\end{table*}

We also manually reviewed each of the unbiased natural language orders from the free-style collection task, and removed 275 utterances requesting toppings or drink items too unrealistic to appear on an actual pizza restaurant menu---see Table \ref{table:exgen} for examples. After removing orders without inter-annotator agreement and orders with questionable semantics (\textit{a pie with peppers, ham and extra ham}) or with non-projective semantics (since these cannot be represented in TOP), we ended up with 441 annotated utterances.
\begin{table*}[h!]
  \centering
  {\small 
	\caption{Subset of utterances obtained from the unconstrained Mechanical Turk generation task.} \label{table:exgen}
  		\smallskip
	\begin{tabular}{|l||c|c|}
		\hline
		\textbf{Unconstrained human generated order} & 	\textbf{Valid} & 	\textbf{Difficulty} \\	\hline
		Two bacon and olive pizzas. & Yes & Easy \\ \hline
		\makecell[l]{Put an order in for me that includes two medium pizzas with \\ extra sauce but light cheese and include one sausage pizza.} & Yes & Interm \\ \hline
		\makecell[l]{I want to order two large pizzas with ham and pepperoni,  \\  and with mushrooms on one of them, and six large cokes.} & Yes & Hard \\  \hline
		\makecell[l]{I want to order six extra large pizzas with pepperoni, sausage, \\   and mushrooms, and four extra large pizzas with the same toppings plus \\  green peppers and extra cheese.} & Yes & Hard \\ \hline
		\makecell[l]{I need one large pizza with every topping plus extra black olives on half of it, \\ and a medium pizza with pepperoni and extra cheese, and six large cokes.} & Yes & Hard \\ \hline
		I need a large  ham and onion pizza and  a large order of  spaghetti & \textbf{No} & N/A \\ \hline
		Order me two large pizzas with green peppers and white onions with two sodas. & \textbf{No} & N/A \\ \hline
		I would like to order two sodas and a medium pizza with olives and onions on it. & \textbf{No} &  N/A \\ \hline
		Can I  get a pineapple pizza  on a thin  crust  cut into  squares & \textbf{No} & N/A \\ \hline
		Five medium pizzas with peppers and lobster chunks. & \textbf{No} & N/A \\ \hline
		\makecell[l]{I would like a medium thin crust pizza with with spicy chorizo, Manchego cheese, \\ and red onion, a bottle of Mondavi merlot, and another bottle of Prosecco.} & \textbf{No} & N/A \\ \hline
	\end{tabular}
        }
\end{table*}

The released data received no processing except for removing non-ASCII characters, the sole exception being the \texttt{ñ} in \texttt{jalapeño}. Examples of actual collected utterances and their extracted semantics can be found in Appendix~\ref{app:ex}, Table~\ref{table:exmap}.

\section{Model details}\label{app:model}
\paragraph{Tokenizers}
For LSTMs and RNNGs, we split on whitespace to get independent tokens. 
For our BART-based models we use the GPT-2 \cite{radford2019language} tokenizer prior to feeding natural language utterances into the encoder. Our Insertion Transformer models make use of BPE \cite{sennrich-etal-2016-neural} tokenizers with a vocabulary of $30,000$ to align with the pretrained encoders used in the network. 

\paragraph{Data preprocessing}
We preprocess the dataset before training as follows: (a) we remove the leading \mtt{ORDER} constructor from the target output sequences, since it is a universal top-level constructor and there is nothing to be learned for it; (b) we downcase the entity tokens to facilitate their handling by the tokenizers. 

\paragraph{Pipeline model} The data used in the pipeline system is prepared by converting the strings in TOP format to flattened labels for both the IS and NER model. For example, if the TOP string is:

\smallskip

{\footnotesize
\begin{tabular}{l}
	\hspace*{-0.15in}  \lp{\tt ORDER} {\em I'd like} \lp{\tt PIZZAORDER} {\tt (NUMBER} {\em two}\rp \\
	\hspace*{-0.15in} {\tt (SIZE} {\em large\/}\rp {\em pizzas with\/} {\tt (TOPPING} {\em ham\/}\rp\rp\rp
\end{tabular}
}

the IS labeling will be: \\[0.05in]
\begin{minipage}{3in}
{\footnotesize
\begin{verbatim}
Other Other B-PIZZAORDER I-PIZZAORDER
I-PIZZAORDER I-PIZZAORDER I-PIZZAORDER
\end{verbatim}}
\end{minipage}\\[0.05in]
and the NER labeling will be: \\[0.05in]
\begin{minipage}{3in}
{\footnotesize
\begin{verbatim}
B-NUMBER B-SIZE Other Other B-TOPPING
\end{verbatim}}
\end{minipage}\\[0.05in]
During evaluation, the output from IS and NER model is converted back to TOP or EXR, which is then used to compute the EM score. \\
The models are fine-tuned on the PIZZA dataset using the ADAM optimizer with a Noam scheduler, setting the base learning rate to 0.075. We make use of F1 scores to select the best models for both IS and NER during the validation phase. We observe that the IS model's F1 ranges from 76-79 (after experimenting with different seeds) while the NER model's F1 ranges from 95-96. Intent segmentation is therefore the bottleneck in the pipeline system. This large difference in performance is likely related to the difference in the lengths of the input sequences for the two models. The NER model receives much shorter sequences than the IS model, which should make it less error prone.

There are cases when the IS and NER models can produce solitary \textit{I-} labels. For example, for the utterance {\em two large pizzas with ham\/}, the NER model might output:
{\tt\small B-NUMBER B-SIZE Other Other I-TOPPING}.
Here, the \smtt{I-TOPPING} label is not preceded by \smtt{B-TOPPING}. During evaluation we apply a post-processing step to convert such solitary \textit{I-} labels to \smtt{Other}. Without this step, the model's EM is 62.96 and 19.91 on the test data and on the subset of test data where PCFG makes an error, respectively (with EXR ground truth). 

\paragraph{Hyperparameter tuning}
A total of 5 training and evaluation runs were run for each model.
Final hyperparameter selection was performed on Tesla V100 GPUs by manually tuning and choosing the highest EM during evaluation on the dev set
for all models except for the pipeline model, which used F1 metric.  The best hyperparameter configuration values are in bold,
as follows.

BART searched the following hyperparameter ranges:  Batch size: [\textbf{512}, 768, 1024]; pretrained blocks dropout: [0.05, \textbf{0.10}, 0.20, 0.30]; 
learning rates: [\textbf{0.15}, 0.30, 0.5]; learning rate multiplier schedule: [0 to 1.0 + incremental step every 20k updates, \textbf{0 to 0.0175 +
incremental step every 20k updates}, 0 to 0.0175 + incremental step every 200 updates].  30 search trials were performed.

Insertion Transformer models searched these ranges: Batch size [128, 256, 512], final fine-tuning learning rate multiplier:  [0.002, 0.1, \textbf{0.2}, 0.7]; 
learning rate: [0.05, 0.1, \textbf{0.15}, 0.3]; Dropout: [0.05, \textbf{0.1}, 0.3]. Best batch sizes were 128 (decode-from-source) and 256 (decode-from-scratch).
12 search trials were performed for each decoding strategy model.

RNNG model search included: Beam size: [1, \textbf{5}, 10]; Optimizer: [SimpleSDG, \textbf{Adam}]; Dropout: [0, \textbf{0.5}]; Learning rate decay [0.05, \textbf{0.5}].
Best learning rate schedule: initial rate: 0.001, learning rate decay factor: 0.5, decay interval: 1000.  10 search trials were performed.

Pipeline model search was performed on the Intent Segmentation model; the obtained hyperparameters values were also used on the NER model.
The search range included:
Batch Size: [128, 256, \textbf{512}] examples; (Base learning rate, warm up steps): [\textbf{(0.075, 500)}, (0.100, 500), (0.125, 500), (0.050, 500), (0.075, 250), (0.075, 1000)]; 
k-best viterbi: [\textbf{1},3,5]; ([steps, LR multiplier], final LR multiplier): [([(100, 0.0)], 0.20), ([(100, 0.0), (200, 0.1)], 0.20), \textbf{([(50, 0.0)], 0.20)}, ([(50, 0.0)], 0.30)]. 
16 search trials were performed.

The bi-LSTM was used as a simple baseline for the other systems and hyperparameter search for it was simplified to: (a) enabling decoder attention [\textbf{yes}, no] and (b) learning Rate: [1E-3,\textbf{1E-4}]. The final configuration was an 1-layer 512-dimensional bidirectional-LSTM with decoder attention, trained with the Adam optimizer and using a learning rate of 1E-4, batch size of 1,000 words per batch, and norm-clipping of 0.1.

\section{Entity resolution (ER)} \label{app:topperfectER}
Models trained to predict TOP or TOP-Decoupled notation against EXR ground truth require an additional entity
resolution step to compare against the final EXR. The ER step was performed by a simple rule-based system that maps relevant phrases to unique entity identifiers using the grammar catalogs. In addition, the ER step adds a default {\tt (NUMBER 1)} slot if needed, because the EXR ground truth includes the slot {\tt NUMBER} for every order. Note that in a more complicated domain that might involve millions of entities and high degrees of ER ambiguity, the seq2seq system would likely include a reranker taking contextual signals such as popularity counts as additional inputs; as mentioned in Appendix~\ref{app:gram}, the PCFG is already equipped to handle this scenario. 

The following two studies investigate the role ER plays in the performance of the TOP and TOP-Decoupled models against EXR ground truth in this dataset. 

\paragraph{Results on TOP ground truth}\label{app:topresults}
Table~\ref{table:topres} shows EM results for models trained on TOP and TOP-Decoupled notation against TOP and EXR ground truth.
Observe that performance against TOP ground truth is very close to performance against EXR ground truth, indicating that the errors in TOP and TOP-Decoupled models
are not primarily due to the entity resolution step, as a similar number of errors occur against TOP ground truth.

Interestingly, for the LSTM models, EM against EXR ground truth is actually slightly better than it is against TOP.
One might expect that the ER step can only add errors, not make corrections.  However, there are cases
where the TOP prediction is incorrect but it is salvaged by the ER step, which ultimately produces a correct EXR form.
Here is an example: {\em i want an order of one large pizza}. The {\tt NUMBER} slot is based on
the wrong utterance token in the TOP prediction, but it resolves to the same correct clause, \mtt{(NUMBER 1)}.

\smallskip

{\small
\begin{tabular}{l}
  TOP Prediction: \\
  {\tt (ORDER} {\em i want} {\tt (PIZZAORDER} {\tt (NUMBER} {\em an}\rp \\
  {\em order of one} {\tt (SIZE} {\em large}\rp {\em pizza} {\tt ))} \\
  TOP Ground Truth: \\
  {\tt (ORDER} {\em i want an order of } {\tt (PIZZAORDER} \\
  {\tt (NUMBER} {\em one}\rp {\tt (SIZE} {\em large}\rp {\em pizza} {\tt ))} \\
  TOP Prediction after ER: \\
  {\tt (ORDER (PIZZAORDER (NUMBER 1)} \\
  {\tt (SIZE LARGE)))} \\
  EXR Ground Truth: \\
  {\tt (ORDER (PIZZAORDER (NUMBER 1)} \\
  {\tt (SIZE LARGE)))}
\end{tabular}}

\begin{center}
	\begin{table*}[!hbt]
	{\small
	\centering
	\caption{Test set EM against TOP and EXR ground truth, grouped by notation type used for training. Mean and standard error from
		models trained on 5 different seeds.}\label{table:topres}        
	\smallskip
	\begin{center}
		\begin{tabular}[c]{lccc}
			\hline
			\textbf{Model} & \thead{\textbf{EXR} \\ \textbf{Ground Truth}} & \thead{\textbf{TOP} \\ \textbf{Ground Truth}} \\
			\hline
			\multicolumn{1}{l}{\em TOP\/}         & & & \\
			\hspace{3mm}Insert-Ptr-Decode-Source          & 40.13 \texttt{\textpm} 0.50 & 40.82 \texttt{\textpm} 0.51 \\
			\hspace{3mm}LSTM                              & 41.44 \texttt{\textpm} 2.15 & 40.44 \texttt{\textpm} 2.08 \\
			\hspace{3mm}Insert-Ptr-Decode-Scratch         & 42.05 \texttt{\textpm} 0.58 & 42.65 \texttt{\textpm} 0.61 \\
			\hspace{3mm}RNNG                              & 50.27 \texttt{\textpm} 3.52 & 50.95 \texttt{\textpm} 3.59 \\
			\hspace{3mm}BART                              & 62.17 \texttt{\textpm} 0.31 & 62.45 \texttt{\textpm} 0.34 \\
			\hspace{3mm}Pipeline                          & 64.08 \texttt{\textpm} 0.32 & 65.22 \texttt{\textpm} 0.32 \\
			\multicolumn{1}{l}{\em TOP-Decoupled\/}& & & \\
			\hspace{3mm}LSTM                              & 42.15 \texttt{\textpm} 1.52 & 41.18 \texttt{\textpm} 1.50 \\
			\hspace{3mm}Insert-Ptr-Decode-Scratch         & 47.89 \texttt{\textpm} 0.21 & 47.89 \texttt{\textpm} 0.21 \\
			\hspace{3mm}BART                              & 71.73 \texttt{\textpm} 0.13 & 71.79 \texttt{\textpm} 0.10 \\
			\hline
		\end{tabular}
	\end{center}
	}
	\end{table*}
\end{center}

\paragraph{Results with perfect ER}
In the PIZZA dataset, there are certain entities which appear only during test/validation, but not during training. If we were to have a record of all possible entities and their aliases as they appear in the entire dataset (including the dev and test sets), then ER from the TOP variants to EXR could improve. We verified this by adding all entities into our ER pipeline and running BART models on the TOP variants. Table \ref{table:BARTperfectER} shows that EM improves by 9.5\% and 4.5\% on the PCFG error test subset when we use the TOP and TOP-Decoupled variants respectively.
Total performance improvements on the overall test set are much more modest, showing that even with perfect ER, the performance of the TOP variants remains significantly lower than the performance of the model that directly generates EXR. 

\begin{center}
	\begin{table*}[tp]
	{\small
	\centering
	\caption{EM against EXR ground truth - with the original grammar entities and all entities.}\label{table:BARTperfectER}
	\smallskip
	\begin{center}
		\begin{tabular}[c]{lccc}
			\hline
			\textbf{Model} & \thead{\textbf{PCFG Error} \\ \textbf{Test Subset}} & \thead{\textbf{Test}} \\
			\hline
			\multicolumn{1}{l}{\em TOP\/}         & & & \\
			\hspace{3mm}BART (original)          & 27.56 \texttt{\textpm} 0.31 & 62.17 \texttt{\textpm} 0.31 \\
			\hspace{3mm} BART (all entities)                              & \textbf{30.18}\texttt{\textpm} \textbf{0.28} & \textbf{62.64 }\texttt{\textpm} \textbf{0.30} \\
			\multicolumn{1}{l}{\em TOP-Decoupled\/}& & & \\
			\hspace{3mm}BART (original)                              & 40.55 \texttt{\textpm} 0.50 & 71.73 \texttt{\textpm} 0.13 \\
			\hspace{3mm}BART (all entities)                             & \textbf{42.54} \texttt{\textpm} \textbf{0.53} & \textbf{72.44} \texttt{\textpm} \textbf{0.13} \\
			\hline
		\end{tabular}
	\end{center}
        }
	\end{table*}
\end{center}

\section{Clause permutation experiments}
\label{App:PermExps}

To further investigate the impact of concrete semantic representation choices on model performance, we analyzed how rearranging the order of sibling nodes in the linearized EXR affects training. For a given input utterance, such as {\em two pizzas with no cheese and a bottle of seven up}, there is a large number of equivalent linearized EXRs with the same semantics (this number grows exponentially with the size of the utterance).

All three entries in Table \ref{table:permutationExp} have identical semantics, but the EXR-natural order is closely aligned with the order in which the entities of interest appear in the natural language utterance. When reading the utterance and the ``natural'' EXR simultaneously from left to right, we see the same entities appearing in the same order: a pizza order, information about number, topping, then a drink order, and information about container and drink type.

\begin{table*}[bp] 
  \centering
  {\small 
	\caption{Different ways of representing the linearized semantic EXR for the same input. Aligning source and target order eases training.}\label{table:permutationExp}
	\smallskip
	\begin{tabular}{|l||l|}
		\hline
		\makecell[l]{Natural Language} &  two pizzas with no cheese and a bottle of seven up  \\	\hline
		\makecell[l]{EXR-natural order}  & \makecell[l]{\smtt{(ORDER} \smtt{(PIZZAORDER} \smtt{(NUMBER 2)} \smtt{(NOT } \smtt{(TOPPING CHEESE)))} \\ \smtt{(DRINKORDER} \smtt{(CONTAINERTYPE BOTTLE )} \smtt{(DRINKTYPE SEVEN\_UP)))}}  \\ \hline
		\makecell[l]{EXR-random  order }  & \makecell[l]{\smtt{(ORDER} \smtt{(DRINKORDER}  \smtt{(DRINKTYPE SEVEN\_UP)}  \smtt{(CONTAINERTYPE BOTTLE ))}\\\smtt{(PIZZAORDER}  \smtt{(NUMBER 2)} \smtt{(NOT } \smtt{(TOPPING CHEESE))))}}   \\ \hline
		\makecell[l]{EXR-sorted string order }  & \makecell[l]{\smtt{(ORDER} \smtt{(DRINKORDER}  \smtt{(CONTAINERTYPE BOTTLE )} \smtt{(DRINKTYPE SEVEN\_UP))} \\ \smtt{(PIZZAORDER} \smtt{(NOT } \smtt{(TOPPING CHEESE))} \smtt{(NUMBER 2)))}}  \\ \hline
	\end{tabular}
        }
\end{table*}

The EXR-random order corresponds to a random permutation of siblings at each level of the tree. In the example given in Table~\ref{table:permutationExp}, the \mtt{PIZZAORDER} and \mtt{DRINKORDER} siblings are switched, as well as the  \mtt{DRINKTYPE} and  \mtt{CONTAINERTYPE} siblings, but not  \mtt{NUMBER} and \mtt{NOT}.

Finally, the EXR-sorted string order is obtained by applying a deterministic ordering function at each level of the tree, which is a simple lexicographic sort over a list of strings. For example, when given the list [\mtt{PIZZAORDER}, \mtt{DRINKORDER}], the returned order is [\mtt{DRINKORDER}, \mtt{PIZZAORDER}] since 'D' precedes 'P' in the alphabet. Similarly,  \mtt{NUMBER} and
\mtt{NOT} are reversed since \mtt{NO} precedes \mtt{NU} in alphabetical order.

The order of siblings in the experiments reported in this paper is almost identical to the natural order, in that the ordering of each entity among each suborder (pizza or drink) is the natural order, but the ordering across the higher level suborders is reversed. To illustrate, if the natural language specifies a pizza order and then a drink order as in the above example, then the linearized EXR representation will be of the form \mtt{(ORDER} \mtt{(DRINKORDER ..)} \mtt{(PIZZAORDER ..)}

We trained our models with 3 different orderings in training EXRs:
\begin{itemize}
	\item the reference setting with the ordering used to report results throughout this paper as described above;
	\item a randomly permuted sibling setting; and 
	\item the string list sorting ordering setting. 
\end{itemize}

To better understand the significance of any discrepancies observed among these three settings, we ran a sanity experiment where the setting is the same as in the reference but with a changed training seed. That allowed us to determine whether changes due to a different ordering are significant relative to changes as simple as changing the seed.

\medskip

The dev unordered EM over the course of training can be observed in Figure~\ref{fig:permutation}. We can make the following observations:

\begin{figure*}[h!]
	\centering
	\includegraphics[width=\textwidth]{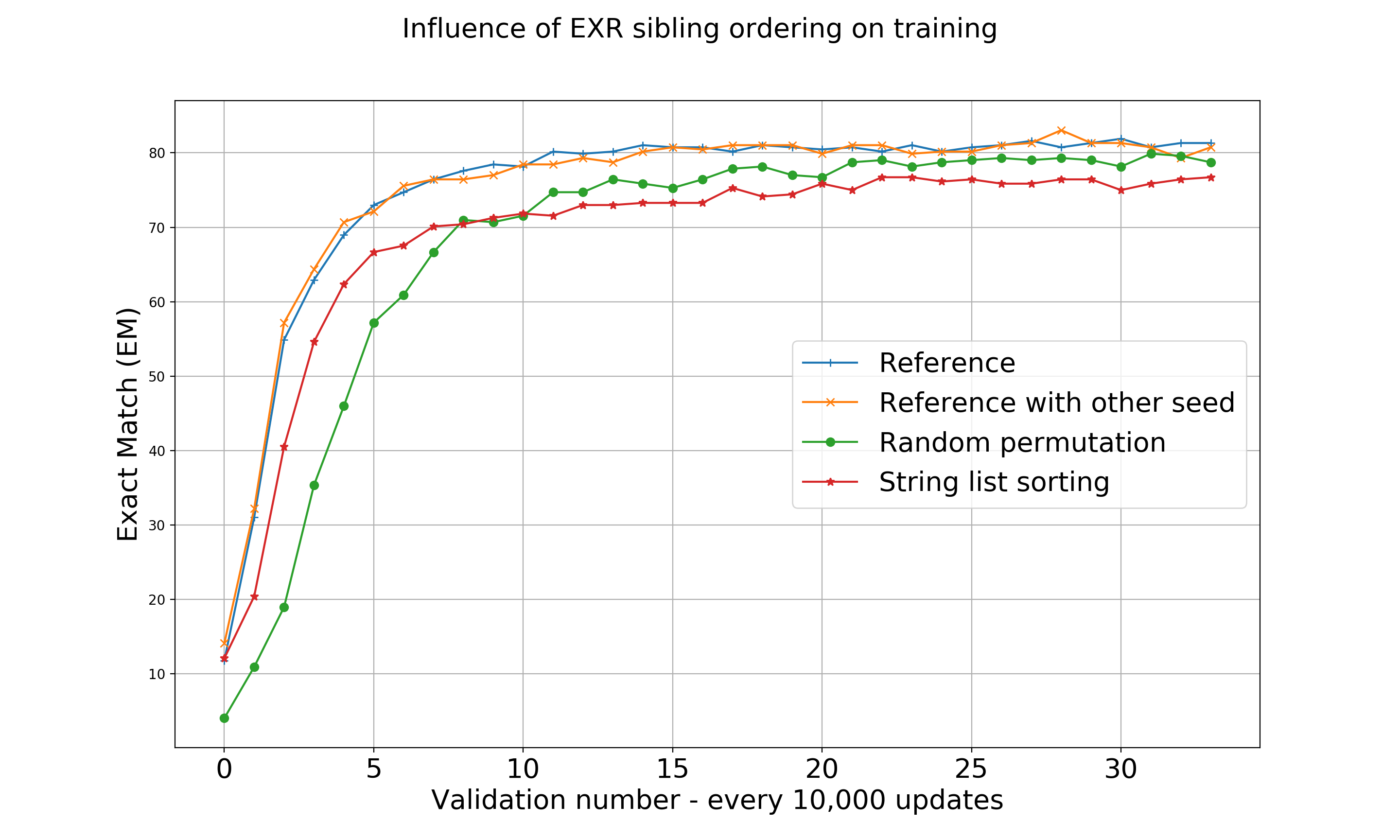}
	\caption{Dev performance along training when using different ordering of clauses in linearized EXR}
	\label{fig:permutation}
\end{figure*}

\begin{itemize}
	\item A clause ordering that is more aligned with the given utterance eases training, as the blue and orange curves reach better performance much faster than the red and green curves. 
	\item The random permutation (green curve) is not as bad as the deterministic reordering (red curve). Eventually it catches up to the blue curve, indicating that the model learned to disregard the noise we artificially injected by reordering the nodes.
	\item The lexicographic ordering (red curve) leads to faster gains initially but ends up much lower than the random one. We hypothesize than learning the artificial but systemic ordering is easier for the model initially than trying to figure out the order pattern, but is detrimental in the long run.
	\item All of the foregoing observations are significant compared to the much smaller difference brought about by simply changing the training seed (blue and orange curves).
\end{itemize}

\section{Training data ablations for BART on EXR} \label{app:ablation}
\begin{figure}[b!]
	\centering
	\includegraphics[width=0.5\textwidth]{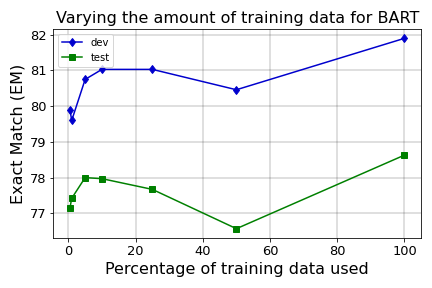}
	\caption{Ablation study on BART by varying the amount of training data}
	\label{fig:ablation}
\end{figure}

We performed ablation studies with our best BART model by varying the amount of training data fed to the model. Figure~\ref{fig:ablation} shows our results when we vary the training data size from 0.5\% ($\sim$12k samples) to 100\% ($\sim$2 million samples).

We observed that reducing the amount of training data results in BART models that perform 
as well as the BART models that are trained on the full set. This is due primarily to the large
amount of redundancy in the large training set, and also to BART's pretraining. We noticed that
the models take longer to converge when using less training data. For instance, the 0.5\% model
had to be trained for 200 epochs,  whereas the final 100\% model converges within one single epoch.

\section{Data examples}\label{app:ex}
While the full dataset will be made available through an external repository,  the following short list gives a flavor of these utterances:
\begin{itemize}
\item {\em one pie with ham\/} is an easy synthetically generated order;
  \item {\em can I get one medium size with extra bacon and mushrooms but little cheese and a small pepsi\/} is a harder synthetically generated order;
  \item {\em today I'll get a ham and bacon pizza, medium size} is an easy human-generated order; and
  \item {\em two pepperoni pies, one with mushrooms and the other with cheese on half only\/} is a harder human-generated order.
\end{itemize}    

Table~\ref{table:exmap} gives additional examples of natural language orders and their EXR semantics. The examples in that table are the result of the two
Mechanical Turk tasks described in Section~\ref{Sec:Dataset} and Appendix~\ref{App:DataAndModelDetails}.

\begin{table*}[!hbt]
  \centering {\small   
	\caption{Some utterances obtained from Mechanical Turk tasks.} \label{table:exmap}
	\smallskip
	\begin{tabular}{|l||l|}
		\hline
		\textbf{Natural Language} & 	\textbf{EXR Linearized Semantics} \\	\hline
		\makecell[l]{I like to have two medium size pizzas} & \smtt{(ORDER} \smtt{(PIZZAORDER} \smtt{(NUMBER 2)} \smtt{(SIZE MEDIUM))) } \\ \hline
		\makecell[l]{I want a vegan pizza, small.} & \makecell[l]{\smtt{(ORDER} \smtt{(PIZZAORDER} \smtt{(NUMBER 1)} \smtt{(SIZE SMALL)}\\ \smtt{(STYLE VEGAN)))}}  \\  \hline
		\makecell[l]{I need to put in an order of two large \\ cheese pizzas with extra cheese } & \makecell[l]{\smtt{(ORDER} \smtt{(PIZZAORDER} \smtt{(NUMBER 2)} \smtt{(SIZE LARGE)} \\ \smtt{(COMPLEX\_TOPPING} \smtt{(QUANTITY EXTRA)} \\ \smtt{(TOPPING CHEESE)))} } \\ \hline
		\makecell[l]{i'll try two medium pizzas and tuna \\ with extra cheese but hold on pesto} &	\makecell[l]{\smtt{(ORDER} \smtt{(PIZZAORDER} \smtt{(NUMBER 2)} \smtt{(SIZE MEDIUM)}\\\smtt{(COMPLEX\_TOPPING} \smtt{(QUANTITY EXTRA)} \\ \smtt{(TOPPING CHEESE))} \smtt{(NOT} \smtt{(TOPPING PESTO))} \\ \smtt{(TOPPING TUNA)))}} \\ \hline
		\makecell[l]{Get me a pineapple and bacon pizza \\ without the extra cheese.} &	\makecell[l]{\smtt{(ORDER} \smtt{(PIZZAORDER} \smtt{(NUMBER 1)} \\ \smtt{(TOPPING PINEAPPLE)} \smtt{(TOPPING BACON)} \\ \smtt{(NOT} \smtt{(COMPLEX\_TOPPING} \smtt{(QUANTITY EXTRA)} \\ \smtt{(TOPPING CHEESE)))} } \\ \hline
		\makecell[l]{Two large pizzas, one with \\ cherry tomatoes and one with onions} & \makecell[l]{\smtt{(ORDER} \smtt{(PIZZAORDER} \smtt{(NUMBER 1)} \smtt{(SIZE LARGE)} \\ \smtt{(TOPPING CHERRY\_TOMATOES))} \smtt{(PIZZAORDER} \\ \smtt{(NUMBER 1)} \smtt{(SIZE LARGE)} \smtt{(TOPPING ONIONS)))} } \\ \hline
		\makecell[l]{Can I  have a  small pizza  with salami and \\ red onion with a  little basil on  the  top} &	 \makecell[l]{\smtt{(ORDER} \smtt{(PIZZAORDER} \smtt{(NUMBER 1)} \smtt{(SIZE SMALL)} \\ \smtt{(COMPLEX\_TOPPING} \smtt{(QUANTITY LIGHT)} \\ \smtt{(TOPPING BASIL))} \smtt{(TOPPING RED\_ONIONS)} \\ \smtt{(TOPPING SALAMI)))}}  \\ \hline
		\makecell[l]{I want to order a medium big meat pizza \\ and two medium cokes.} & \makecell[l]{\smtt{(ORDER} \smtt{(DRINKORDER} \smtt{(DRINKTYPE COKE)}\\ \smtt{(NUMBER 2)} \smtt{(SIZE MEDIUM))} \smtt{(PIZZAORDER} \smtt{(NUMBER 1)} \\ \smtt{(SIZE MEDIUM)} \smtt{(STYLE MEAT\_LOVER)))} } \\ \hline
		\makecell[l]{Put in an order for two cans of coke and one \\ large pizza with beef and black olives.} &  \makecell[l]{\smtt{(ORDER} \smtt{(DRINKORDER} \smtt{(CONTAINERTYPE can)} \\ \smtt{(DRINKTYPE COKE)} \smtt{(NUMBER 2))} \smtt{(PIZZAORDER} \\ \smtt{(NUMBER 1)} \smtt{(SIZE LARGE)} \smtt{(TOPPING BEEF)} \\ \smtt{(TOPPING OLIVES)))} } \\ \hline
		\makecell[l]{I'd like a small supreme pizza and a \\ two liter coke.} &	 \makecell[l]{\smtt{(ORDER} \smtt{(DRINKORDER} \smtt{(DRINKTYPE COKE)} \\ \smtt{(NUMBER 1)} \smtt{(VOLUME 2 LITER))} \smtt{(PIZZAORDER} \\ \smtt{(NUMBER 1)} \smtt{(SIZE SMALL)} \smtt{(STYLE SUPREME)))} } \\ \hline
		\makecell[l]{I need twenty pizzas for my party, make \\ half of them pepperoni and the \\ other ten cheese.} &	 \makecell[l]{\smtt{(ORDER} \smtt{(PIZZAORDER} \smtt{(NUMBER 10)} \smtt{(TOPPING CHEESE))} \\ \smtt{(PIZZAORDER} \smtt{(NUMBER 10)} \smtt{(TOPPING PEPPERONI)))}}  \\ \hline
		\makecell[l]{i want a couple of pizzas; make them both \\ medium please. i want mushrooms, pesto \\ but hold the onions.} & \makecell[l]{\smtt{(ORDER} \smtt{(PIZZAORDER} \smtt{(NUMBER 2)} \smtt{(SIZE MEDIUM)} \\ \smtt{(TOPPING PESTO)}  \\ \smtt{(NOT} \smtt{(TOPPING ONIONS))} \smtt{(TOPPING MUSHROOMS)))} } \\ \hline
		\makecell[l]{Order me two large cheese pizzas one with \\ pineapple and the other with jalapenos.} &	 \makecell[l]{\smtt{(ORDER} \smtt{(PIZZAORDER} \smtt{(NUMBER 1)} \smtt{(SIZE LARGE)} \\ \smtt{(TOPPING CHEESE)} \smtt{(TOPPING JALAPENO\_PEPPERS))} \\ \smtt{(PIZZAORDER} \smtt{(NUMBER 1)} \smtt{(SIZE LARGE)} \\ \smtt{(TOPPING CHEESE)} \smtt{(TOPPING PINEAPPLE)))}} \\ \hline
	\end{tabular}
}        

\end{table*}

As Table~\ref{table:exgen} illustrates, the obtained orders came with varying degrees of sophistication and some of the collected responses are quite challenging to parse correctly. The task also resulted in a number of invalid orders that we filtered out as described in Appendix~\ref{App:DataAndModelDetails}.

\section{Other models}\label{app:app_other_models}

\subsection{Insertion based transformers}

In addition to auto-regressive decoding, we 
explored the non-auto-regressive generation techniques introduced in
Insertion Transformers \cite{stern2019insertion}, using utterances as 
the source and TOP representations as the target. We followed the 
parallel generation strategy where more than a single token can be generated 
at any time step. We explored decoding from scratch (inserting both semantic labels and utterance tokens) as well as
decoding from source (inserting only semantic tokens around the full source utterance sequence initialized at start).
We used a pointer generator network  \cite{see-etal-2017-get} in conjunction with a pretrained 12-layer
encoder and 4-layer decoder in all our experiments. Because these models are pointer generators, where the leaf node
slots are generated by pointing back to the utterance input, they cannot be trained on EXR notation.
The decode-from-source model requires the entire utterance sequence to be present
in the target, so it also cannot be trained on TOP-Decoupled notation.
The models were fine-tuned using validation-based early stopping
with a learning rate of 0.15.

The poor results observed in Table~\ref{table:exrresApp} could be explained by a smaller 4-layer decoder, a smaller pre-training dataset (Wikipedia only) and lack of beam search during decoding.

\subsection{RNNGs}
We also explored recurrent neural network grammars (RNNGs) and implemented beam search on top of the framework provided
by~\cite{dyer2016recurrent}. We used a beam size of $5$ and preserved the rest of the settings as given by \cite{dyer2016recurrent}.
Here the source remains the natural language utterance and the target is a TOP tree produced by a sequence of SHIFT,
REDUCE, and NT (non-terminal) actions. The model is trained to predict the sequence of actions that assemble the target parse tree.
Because this sequence of actions is applied against the entire source utterance, this RNNG model cannot be trained on EXR or
TOP-Decoupled notations.

As shown in Table~\ref{table:exrresApp} the RNNG results are not competitive with seq2seq approaches, as initially found out in~\cite{rongali2020don}.

\begin{center}
	\begin{table*}[!hbt]
		{\small
			\centering
			\caption{EM against EXR ground truth, grouped by notation type used for training. Mean and standard error from
	models trained on 5 different seeds.}\label{table:exrresApp}
			\smallskip
			\begin{center}
				\begin{tabular}[c]{lccc}
					\hline
					\textbf{Model} & \thead{\textbf{Dev} \\ \textbf{(348 utt)}} & \thead{\textbf{Test} \\ \textbf{(1357 utt)}}  &  \thead{\textbf{PCFG Error} \\ \textbf{Test Subset} \\ \textbf{(434 utt)}} \\
					\hline
					PCFG Parser                           & 69.54 & 68.02 & 0.0  \\ \hline
					\multicolumn{1}{l}{\em TOP\/}         & & & \\
					\hspace{3mm}Insert-Ptr-Decode-Source          & 40.00 \texttt{\textpm} 0.83 & 40.13 \texttt{\textpm} 0.50 & 8.85 \texttt{\textpm} 0.38 \\
					\hspace{3mm}Insert-Ptr-Decode-Scratch         & 45.29 \texttt{\textpm} 0.45 & 42.05 \texttt{\textpm} 0.58 & 10.78 \texttt{\textpm} 0.30 \\
					\hspace{3mm}RNNG                             & 55.06 \texttt{\textpm} 3.50 & 50.27 \texttt{\textpm} 3.52 & 20.83 \texttt{\textpm} 3.86 \\
					\multicolumn{1}{l}{\em TOP-Decoupled\/}& & & \\
					\hspace{3mm}Insert-Ptr-Decode-Scratch         & 50.69 \texttt{\textpm} 0.19 & 47.89 \texttt{\textpm} 0.21 & 13.78 \texttt{\textpm} 0.45 \\

					\hline
				\end{tabular}
			\end{center}
		
	}
\end{table*}
\end{center}

\end{document}